%% file: sample-sigconf.tex
\documentclass[sigconf]{acmart}
\usepackage{booktabs, makecell, multirow, tabularx, balance} 
\AtBeginDocument{%
  \providecommand\BibTeX{{%
    \normalfont B\kern-0.5em{\scshape i\kern-0.25em b}\kern-0.8em\TeX}}}

\setcopyright{acmcopyright}
\copyrightyear{2023}
\acmYear{2023}
\setcopyright{acmlicensed}\acmConference[CIKM '23]{Proceedings of the 32nd ACM International Conference on Information and Knowledge Management}{October 21--25, 2023}{Birmingham, United Kingdom}
\acmBooktitle{Proceedings of the 32nd ACM International Conference on Information and Knowledge Management (CIKM '23), October 21--25, 2023, Birmingham, United Kingdom}
\acmPrice{15.00}
\acmDOI{10.1145/3583780.3615028}
\acmISBN{979-8-4007-0124-5/23/10}

\begin{document}
\title{Relation-Aware Diffusion Model for Controllable Poster Layout Generation}


\author{Fengheng Li}
\authornote{Both authors contributed equally to this research.}
\authornote{Work done during an internship at JD.com.}

\affiliation{%
   \department{TKLNDST, CS}
   \institution{Nankai University}
   \city{Tianjin}
   \country{China}
}
\email{lifengheng@foxmail.com}

\author{An Liu}
\authornotemark[1]
\affiliation{%
   \institution{Retail Platform Operation and
Marketing Center, JD}
   \city{Beijing}
   \country{China}
}
\email{liuan39@jd.com}

\author{Wei Feng}
\affiliation{%
   \institution{Retail Platform Operation and
Marketing Center, JD}
   \city{Beijing}
   \country{China}
}
\email{	fengwei25@jd.com}

\author{Honghe Zhu}
\affiliation{%
   \institution{Retail Platform Operation and
Marketing Center, JD}
   \city{Beijing}
   \country{China}
}
\email{zhuhonghe1@jd.com}

\author{Yaoyu Li}
\affiliation{%
   \institution{Retail Platform Operation and
Marketing Center, JD}
   \city{Beijing}
   \country{China}
}
\email{	liyaoyu1@jd.com}

\author{Zheng Zhang}
\affiliation{%
   \institution{Retail Platform Operation and
Marketing Center, JD}
   \city{Beijing}
   \country{China}
}
\email{zhangzheng11@jd.com}

\author{Jingjing Lv}
\affiliation{%
   \institution{Retail Platform Operation and
Marketing Center, JD}
   \city{Beijing}
   \country{China}
}
\email{lvjingjing1@jd.com}

\author{Xin Zhu}
\affiliation{%
   \institution{Retail Platform Operation and
Marketing Center, JD}
   \city{Beijing}
   \country{China}
}
\email{zhuxin3@jd.com}

\author{Junjie Shen}
\affiliation{%
   \institution{Retail Platform Operation and
Marketing Center, JD}
   \city{Beijing}
   \country{China}
}
\email{shenjunjie@jd.com}

\author{Zhangang Lin}
\affiliation{%
   \institution{Retail Platform Operation and
Marketing Center, JD}
   \city{Beijing}
   \country{China}
}
\email{	linzhangang@jd.com}

\author{Jingping Shao}
\affiliation{%
   \institution{Retail Platform Operation and
Marketing Center, JD}
   \city{Beijing}
   \country{China}
}
\email{shaojingping@jd.com}

\renewcommand{\shortauthors}{Fengheng Li et al.}



\begin{abstract}
Poster layout is a crucial aspect of poster design. Prior methods primarily focus on the correlation between visual content and graphic elements. However, a pleasant layout should also consider the relationship between visual and textual contents and the relationship between elements. In this study, we introduce a relation-aware diffusion model for poster layout generation that incorporates these two relationships in the generation process. Firstly, we devise a visual-textual relation-aware module that aligns the visual and textual representations across modalities, thereby enhancing the layout's efficacy in conveying textual information. Subsequently, we propose a geometry relation-aware module that learns the geometry relationship between elements by comprehensively considering contextual information. Additionally, the proposed method can generate diverse layouts based on user constraints. To advance research in this field, we have constructed a poster layout dataset named CGL-Dataset V2. Our proposed method outperforms state-of-the-art methods on CGL-Dataset V2. The data and code will be available at https://github.com/liuan0803/RADM.

\end{abstract}






\maketitle

\section{Introduction}
Poster layout generation aims to predict the position and category of graphic elements on the image, which is important for visual aesthetics and information transmission of posters. Due to the need to consider both graphic relationships and image compositions when creating high-quality poster layouts, this challenging task is usually completed by professional designers. 
However, manual design is often time-consuming and financially burdensome.
\input{Figtex/compare1}

To generate high-quality poster layouts at low cost, automatic layout generation has become increasingly popular in academia and industry. With the advent of deep learning, some content-agnostic methods \cite{Jyothi2019LayoutVAESS,BLT2022ECCV, Yang_2021_CVPR,inoue2023layout, Lee2019NeuralDN, hui2023unifying} are proposed to learn the internal relationship of graphic elements.
However, these methods prioritize the graphic relationships between elements and overlook the impact of visual content on poster layout. Therefore, applying these methods directly to poster layout generation can negatively impact subject presentations, text readability and the visual balance of the poster as a whole. To address these issues, several content-aware methods \cite{Li2019LayoutGANGG,Cao2022GeometryAV,Zhou2022CompositionawareGL} generate layouts based on the visual contents of input background images. ContentGAN \cite{Li2019LayoutGANGG} leverages visual and textual semantic information to implicitly model layout structures and design principles, resulting in plausible layouts. However, ContentGAN lacks spatial information. To overcome this limitation, CGL-GAN \cite{Zhou2022CompositionawareGL} combines a multi-scale CNN and a transformer to extract not only global semantics but also spatial information, enabling better learning of the relationship between images and graphic elements.

Despite their promising results, two relationships still require consideration in poster layout generation. On one hand, text plays an important role in the information transmission of posters, so the poster layout generation should also consider the relationship between text and vision. As shown in the first row in Fig.~\ref{fig:compare1}, ignoring text during layout generation will result in the generated
layout not being suitable for filling the given text content. On the other hand, a good layout not only needs to consider the position of individual elements, but also the coordination relationship between elements. As shown in the second row in Fig.~\ref{fig:compare1}, considering the geometric relationships between elements can work better on graphic metrics.

In this paper, we propose a relation-aware diffusion model for poster layout generation as depicted in Fig. \ref{fig:overview}, considering both visual-textual and geometry relationships. As diffusion models have achieved great success in many generation tasks \cite{austin2021structured, Ruiz_2023_CVPR, Blattmann_2023_CVPR, Zhang_2023_CVPR}, we follow the noise-to-layout paradigm to generate poster layout by gradually adjusting noisy layout via the learned denoising model. In each sampling step, given a set of boxes sampled in Gaussian distribution or the estimated boxes from the last sampling step as input, we extract RoI features from the feature map generated by the image encoder. Then a Visual-Textual Relation-Aware Module (VTRAM) is proposed to model the relationship between visual and textual features, which makes the layout result determined by both the image and text content. Meanwhile, we design a Geometry Relation-Aware Module (GRAM) to enhance the features of each RoI based on its relative position to other RoIs. This enables the model to better understand the contextual information of graphic elements. Finally, the position and category of elements are determined by the outputs of VTRAM and GRAM, as well as the RoI features. The predicted results are sent to the next step to progressively refine themselves. Benefiting from the newly proposed VTRAM and GRAM, users can regulate the layout generation process by predefining layouts or adjusting text content.

To summarize, the contributions of our work are listed below:
\begin{itemize}
    \item We propose a novel visual-textual relation-aware module to study the relationship between visual and textual information, which makes the generated layout results easier for posters to convey text information.
    \item A geometry relation-aware module is used to explicitly learn the geometric relationships between elements, so that each element can consider the context more comprehensively.
    \item To promote research in this field, we extend the dataset proposed in CGL-GAN \cite{Zhou2022CompositionawareGL} to CGL-Dataset V2 by adding text content annotations. Extensive experiments show that our method outperforms state-of-the-art methods, and can generate layout based on user constraints.    
\end{itemize}

\section{Related Work}

\subsection{Layout Generation}
In recent years, there has been a surge of interest in the field of layout generation. 
Researchers have been exploring new techniques and algorithms to automate the process of designing layouts for various applications, such as web design \cite{web2011,Pang2016DirectingUA}, graphic design \cite{Cao2012, Zhou2022CompositionawareGL, zheng-sig19}, and even interior design \cite{Yu2011MakeIH}.
Various techniques have been proposed to generate layouts automatically that are visually appealing and semantically meaningful.
Prior approaches can be roughly divided into two subcategories: rule-based and template-based methods.
Rule-based methods \cite{Cao2012, ODonovan2014LearningLF, Pang2016DirectingUA} define a set of rules that govern the placement of various elements in a layout. 
These rules are based on design principles and heuristics that have been established by experts in the field.
Template-based methods \cite{Jacobs2003AdaptiveGD, qian2020retrieve} involve using pre-defined templates to generate layouts that conform to specific design patterns. 
However, the methods mentioned above require professional knowledge and the generated layouts usually lack diversity.
\input{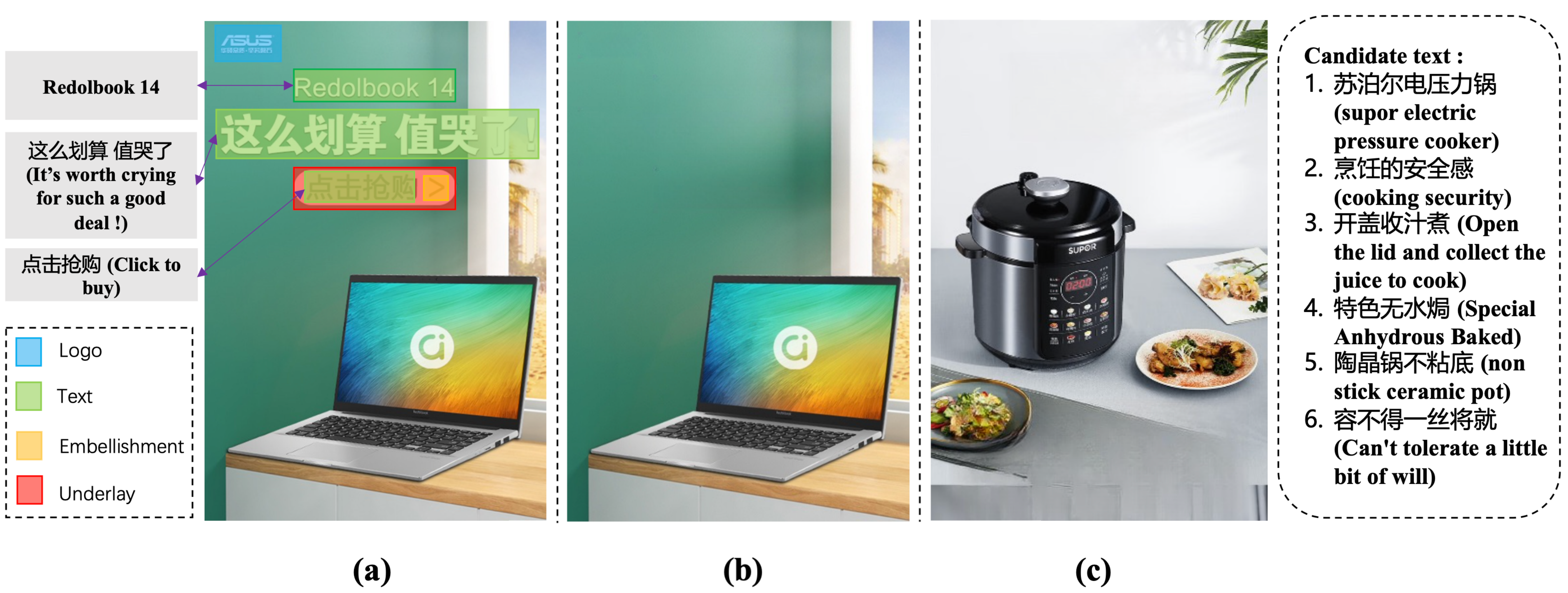}
According to whether the visual content is considered, we divide the deep generative models into two categories: content-agnostic and content-aware methods.
Content-agnostic methods usually yield layouts with visual balance and symmetry as there are fewer constraints, making them suitable for documents, user interfaces, and publication generation. 
LayoutVAE \cite{Jyothi2019LayoutVAESS}, which utilizes Variational Autoencoders, is a method that learns to produce layouts based on the categories of elements. 
To further improve the quality of the generated layouts, transformers \cite{BLT2022ECCV, Yang_2021_CVPR} are used in the generation task.
Due to the attention mechanism, transformer-based methods are capable of implicitly learning the relationships between elements.

Nonetheless, content-agnostic methods tend to have inadequate performance when it comes to layout generation tasks that require comprehension of given content. 
To solve the problem, content-aware methods are proposed for specific tasks. 
ContentGAN \cite{zheng-sig19} is the first model to incorporate both visual and textual semantics in the generation of magazine layouts. It used Generative Adversarial Networks (GANs) to learn complicated layout structures and generate layouts from noise, which enables the diversity of layouts. However, the lack of spatial information and detailed features of the image leads to unsatisfactory layout results under complex background conditions. More recently, transformer-based models such as CGL-GAN \cite{Zhou2022CompositionawareGL} and LCVT \cite{Cao2022GeometryAV} have been introduced for stronger layout capabilities. Although these methods introduce spatial visual information and domain alignment information respectively, they do not consider the impact of text content on layout and how to more accurately model the positional relationship between layout elements. Different from the above methods, we introduce visual and textual prior knowledge to generate layouts and consider geometric relation priors to strengthen the feature expression between layout elements.

\subsection{Diffusion Models}
In recent years, diffusion models \cite{DDIM2020, DDPM2020} have gradually become the focus of generative tasks because of their impressive high-quality generative capabilities. The diffusion and denoising processes are key components of this approach.
Diffusion refers to the gradual transformation of an initial image into a final noisy image through a series of small, random perturbations. 
Denoising, on the other hand, is the process of learning to remove noise from the image to actual distribution. Besides image generation, Diffusion models are gaining momentum in various fields and showing promising performance. 
DiffusionDet \cite{chen2022diffusiondet} is the first to apply diffusion model for the task of object detection. 
InST \cite{Zhang_2023_CVPR} implemented Inversion-Based Style Transfer with Diffusion Models. Video LDM \cite{Blattmann_2023_CVPR} achieved high-resolution video generation by training a diffusion model in a compressed low-dimensional latent space. Naturally, the diffusion model is also introduced into the field of layout generation. LayoutDM \cite{inoue2023layout} uses a discrete diffusion model to predict the attributes of elements like category and position.
LDGM \cite{hui2023unifying} unifies unconditional and conditional generation in a single diffusion model.
But these methods are oblivious to input contents and perform poorly in poster layout generation.
By introducing a multimodal diffusion model, our method can align the image and texts and produce more visually convincing posters.

\section{CGL-Dataset V2}

CGL-Dataset V2 is a dataset for the task of automatic graphic layout design of advertising posters, containing 60,548 training samples and 1035 testing samples. It is an extension of CGL-Dataset \cite{Zhou2022CompositionawareGL}. The original CGL-Dataset contains 4 types of elements: logos, texts, underlays and embellishments as shown in Fig. ~\ref{fig:dataset} (a). Each element consists of category and coordinates information. However, it does not include text content annotations, which have a crucial impact on the layout of posters. As shown in Fig. ~\ref{fig:dataset} (a), to study the influence of content, we supplementally annotate the textual content. In the training set, in order to obtain a clean background image for model training, we use an inpainting model \cite{Suvorov_2022_WACV} to erase layout elements, and the result is shown in Fig. \ref{fig:dataset} (b). The text information is not provided in the test set of the original CGL-Dataset, so we additionally collect 1035 poster images with usable textual descriptions to replace the original test set. 
As shown in Fig. ~\ref{fig:dataset} (c), the collected poster images are processed the same as the training set to get a clean background image. Meanwhile, we collected all the promotional slogans of the current product for analysis of different textual content for poster layout impact. Since the collected text content is more focused on the e-commerce field, we use a pre-trained model based on massive e-commerce text corpus training to extract textual features. The extraction method is detailed in section 4.2. For convenience, we will publish the language model for extracting textual features.

\input{Figtex/overview}
\input{Figtex/diffusion}
\section{Method}
The overview of our method is shown in Fig.~\ref{fig:overview}. The proposed method is composed of four parts: feature extractor, Visual-Textual Relation-Aware Module (VTRAM), Geometry Relation-Aware Module (GRAM) and layout decoder. The feature extractor extracts features from text and images respectively. Then VTRAM models the visual and textual relationship for superior layouts. Meanwhile, GRAM is used to strengthen the ability to express the positional relationship between each RoI feature. Finally, based on the outputs of VTRAM and GRAM, as well as the RoI features, the layout decoder predicts the coordinates and category of elements. Next, we will introduce the process of applying the diffusion mechanism to poster layout generation and the details of the four parts.

\subsection{Poster Layout Generation with Diffusion Model}

Diffusion models are a class of probabilistic generative models that convert noise to a representative data sample by using Markovian chain.
As shown in Fig. \ref{fig:diffusion}, we formulate the poster layout generation problem as a noise-to-layout generative process by gradually adjusting the noise layout with a learned denoising model. The poster layout generated by the diffusion model also includes two processes: the diffusion process and the denoising process. Given a poster layout, we gradually add Gaussian noise to corrupt the deterministic layout result, we call this operation the diffusion process. Instead, given an initial random layout, we obtain the final poster layout by stepwise denoising, which is called the denoising process. Next, we will introduce the diffusion process and the denoising process respectively.

\subsubsection{Diffusion Process}
$x_0$ is a set of layout elements, each element consists of coordinates $(x, y, w, h)$, where $x, y, w, h$ represent the horizontal center, vertical center, width and height of the rectangular box, respectively. 
We get sample data $x_0$ from a true data distribution $q(x)$ and gradually add Gaussian noise to sample data in each step $i$. 
We get a sequence of intermediate samples $x_1,\cdots, x_i, \cdots, x_T$.
The noise is controlled by the variance schedule $\beta (\beta_i \in (0,1))$.
\begin{equation}
\begin{aligned}
q(x_i|x_{i-1}) &= \mathcal N(x_i;\sqrt{1-\beta_i}x_{i-1},\beta_i \mathbf{I}),\\
q(x_{1:T}|x_0) &= \prod_{i=1}^{T}q(x_i|x_{i-1}).
\end{aligned}
\end{equation}
With the nice property found by \cite{DDPM2020}, we can directly sample $x_i$ at any arbitrary time step $i$ as:

\begin{equation}
\begin{aligned}
q(x_{i}|x_0) &= \mathcal N(x_i;\sqrt{\hat{\alpha_i}}x_0, (1-\hat{\alpha_i})\mathbf{I}),\\
\hat{\alpha_i} &= \prod^{i}_{j=1}(1-\beta_j).
\end{aligned}
\end{equation}

\subsubsection{Denoise Process}
These conditional probabilities $q(x_{i-1}|x_i)$, however, are intractable. 
Instead, we train a model $f_\theta(t, x_t, I_{img}, I_{text})$ to approximate the reverse process,
where $I_{img}$ is visual input, $I_{text}$ is textual input, the $f_\theta$ reconstructs $x_0$ from $x_t$, combining visual and textual input.
More specifically, in our work, the $x_0$ is no longer an image but a layout annotation consisting of $N$ bounding boxes.
In inference, starting from random boxes, our model gradually modifies the position and size of boxes until a plausible layout is formed.
\subsection{Feature Extractor}
\subsubsection{Image Encoder}

Given a clean background image, we use ResNet-50 \cite{He2015DeepRL} with the Feature Pyramid Network (FPN) \cite{Lin2016FeaturePN} to extract visual features. ResNet-50 has gained widespread popularity due to its exceptional performance in computer vision. 
Besides, we use FPN to produce multi-scale feature maps $F$, which consist of image features from low level to high level. Based on $F$, we extract RoI features \cite{DBLP:journals/corr/Girshick15} $V$ with proposal $x$ as follows:
\begin{equation}
\begin{aligned}
V = RoI Pooling(F, x),
\end{aligned}
\end{equation}
where the shape of $V$ is $(C, W, H)$. In the training stage, the RoI feature comes from the real layout with Gaussian noise added, and it derives by random layout denoising in the inference stage.

\subsubsection{Text Encoder}
Given all the promotional slogans of the product on a poster, we extract textual features through a pre-trained language model RoBERTa \cite{Liu2019RoBERTaAR}. We note that the product description is not simply repeating the product name, but highlighting the selling points of the product.
For instance, if you want to promote a computer, you describe it as "high CPU performance" without mentioning "computer". Therefore, it is important to narrow the gap between the product description and the product itself. To address the problem, we gathered a vast product corpus of 200 million items from JD.com and adapt the same pretraining strategy which comprises Masked Language Model (MLM), Attribute-Value Prediction (AVP), and Tertiary Category Prediction (TCP) to finetune RoBERTa. For MLM, we randomly mask certain words from the input product title and feed it into the language model. This allows the model to predict the original sentence accurately. AVP and TCP are used to predict the value of a product based on its attribute and tertiary category. AVP is utilized to extract product values from the product description by utilizing product attribute queries. TCP involves the analysis and assessment of product information to determine the appropriate category. In order to let the model perceive the relationship between text length and layout, we supplement textual length embedding as a part of text features. Finally, we fuse the content features and length features of the text by concat operation, as the output of the text encoder, denoted as $L \in \mathbb{R}^{D_n \times d}$. It is worth noting that our method is not limited to Chinese. Migrating to another language only requires replacing the text encoder here.

\input{Figtex/VTRAM}

\subsection{Visual-Textual Relation-Aware Module}
Instead of concatenating visual features and text features directly, we design a visual-textual relation-aware module to align the feature domain of the image and texts.
The module is aware of the relationship between visual and textual elements and makes optimal use of features from both images and texts. 
This allows for a more comprehensive understanding of the content. In order to ensure a constant number of texts, we employ a method of padding additional vectors to reach a fixed number $D_n$. This approach offers the advantage of allowing our model to process texts of varying lengths.

Fig.~\ref{fig:VTRAM} depicts the pipeline of VTRAM, which performs the multi-modal fusion of each RoI features $V_i \in \mathbb{R}^{C \times W \times H}$ and linguistic features $L \in \mathbb{R}^{D_n \times d}$ in two steps. First, to add explicit position information in visual features,  the RoI feature $V_i$ and its corresponding position embedding are concatenated to get the visual position feature $V_{ip}$:
\begin{equation}
\begin{aligned}
V_{ip} &=V_i \bigoplus P_g(G_i),
\end{aligned}
\end{equation}
where the $P_g$ is the project function, $G_i$ is the coordinate of the $i$-th RoI.

Second, we use visual position feature $V_{ip}$ as the query and linguistic feature maps $L$ as the key and value:
\begin{equation}
\begin{aligned}
V_{iq} &= P_{q}(V_{ip}),\\
L_{k} &= P_{k}(L),\\
L_{v} &= P_{v}(L),\\
\end{aligned}
\end{equation}
where the $P_q$, $P_{k}, P_{v}$ are the $1\times1$ convolution function to convert the vectors into proper shape.

We calculate the final multi-modal feature $M_i$ as follows:
\begin{equation}
\begin{aligned}
M_i  &= P_{o}(softmax(\frac{V^T_{iq}L_{k}}{\sqrt{C}})L^T_{v}),\\
\end{aligned}
\end{equation}
where the $P_o$ is also a $1\times1$ convolution function. The multi-modal feature $M_i$ gathers textual information that is closely related to RoI features, making visual features textual-aware.

\input{Figtex/GRAM}

\subsection{Geometry Relation-Aware Module}
We construct RoI features combining the results of the denoising process and image features, but these features of RoI are independent. To strengthen the position-aware relationship between RoI features, we designed Geometry Relation-Aware Module (GRAM) to allow the model to better learn the content information relationship between graph elements. The details are as follows.
Firstly, given $N$ RoIs, the relative position feature $R_{ij}$ of two boxes $l_i$ and $l_j$ $(i,j \in \{1,2,\dots, N\})$ is calculated as :
\begin{equation}
R_{ij} = [~\log(\frac{|x_i-x_j|}{w_j}), ~\log(\frac{|y_i-y_j|}{h_j}), ~\log(\frac{w_i}{w_j}), ~\log(\frac{h_i}{h_j})].
\end{equation}
Then, the 4-dimensional vectors are embedded to geometry weights by sin-cos encoding method \cite{Vaswani2017AttentionIA} as $R_{pij}$.
\begin{equation}
\begin{aligned}
PE_{(pos,2k)} &= \sin(\frac{pos}{10000^{8k/d_h}}), \\
PE_{(pos,2k+1)} &=\cos(\frac{pos}{10000^{8k/d_h}}), \\
    R_{p} &= PE(R),\\
\end{aligned}
\end{equation}
where the $pos$ is the position and $k$ is the dimension. 
The $d_h$ we set in our experiment is $64$.
Finally, the geometry weights are normalized by the softmax function which prunes the weak pairwise relation and focuses more on the strong ones.
\begin{equation}
     W = Softmax( R_{p}).
\end{equation}

What we need to emphasize is that there are different positioning strategies for different types of elements.
The underlay should cover others while the rest elements should avoid overlapping.
Therefore, we use extracted RoI features as element category information.
To merge the position and category information, the extracted visual features $V$ are flattened and transformed to vectors in $d_t$ dimension by project function $P$.
Finally, the visual embeddings multiply the geometry weights to get the final geometry features $T$:
\begin{equation}
T = W \cdot P( (V^\prime)),
\end{equation}
where $V^\prime$ is the flattened form of $V$.
\subsection{Layout Decoder}
Similar to the task of object detection, the layout decoder predicts the category and coordinates of elements based on various types of RoI features. We construct the whole input of the layout decoder by fusing the outputs of VTRAM and GRAM, as well as the RoI features. The above process can be expressed as follows:
\begin{equation}
\begin{aligned}
I_{decoder} = M \bigoplus T \bigoplus V,\\
\end{aligned}
\end{equation}
where $I_{decoder}$ represents the input of layout decoder, $M$ is the output of VTRAM, $T$ is the output of GRAM and $V$ refers to the RoI features. $\bigoplus$ represents the fusion method of features, the concat fusion used here. Then, these fused features are sent to the detection heads of bounding box regression and category prediction respectively to get the final coordinates and categories. Based on the above detection head results, we use box regression and classification losses to narrow the gap between the model's predictions and the ground truth, respectively. Meanwhile, in order to avoid excessive overlap between predicted boxes, we supplement giou loss as a penalty. The final weighted loss function is composed as follows:
\begin{equation}
\begin{aligned}
Loss &= \alpha_{cls}*L_{cls} + \alpha_{L1}*L_{L1} + \alpha_{giou}*L_{giou}, \\
\end{aligned}
\end{equation}
where $L_{cls}$, $L_{L1}$ and $L_{giou}$ respectively adopt focal loss \cite{Lin_2017_ICCV}, L1 loss and generalized IoU loss \cite{Rezatofighi_2019_CVPR}. $\alpha_{cls}$, $\alpha_{L1}$ and $\alpha_{giou}$ are weight coefficients for three different types of losses, which are set to 5, 5, and 1 respectively in this paper.

\input{Figtex/main_result}

\section{Experiment}

In this section, we will compare the performance of our method and the SOTA method from both qualitative and quantitative perspectives.
\input{Table/table1}

\subsection{Implementation Details}
We implement the proposed method using Pytorch \cite{paszke2019pytorch} and set the maximum diffusion step for sampling and denoising to 1000. Our model is trained using the AdamW \cite{loshchilov2017decoupled} optimizer with the initial learning rate as 2.5$\times$$10^{-5}$ and the weight decay as $10^{-4}$. We train the model for 100 epochs with batch size 16 on NVIDIA P40 GPU and the image size is normalized to 384$\times$600 in order to improve training efficiency.

\subsection{Evaluation Metrics}
We follow the evaluation metrics in CGL-GAN \cite{Zhou2022CompositionawareGL}, including three aspects: user study, composition-relevant measures and graphic measures.

For the user study, we randomly select 60 images from the test set and obtain the layout results corresponding to different methods and invite two groups of designers (five professional, twenty novice designers). Every designer needs to judge whether the layout result is qualified and select the best layout result for the same image. We denote the percentage passing the quality standard as $P_{qs}$ and the percentage that hits the best layout as $P_{best}$ ($P^*_{qs}$ and $P^*_{best}$ for the professional group) for each method.

Composition-relevant measures such as \textit{Readability and visual balance} $R_{com}$ and \textit{Presentation of subjects} ($R_{csub}$ and $R_{shm}$) are introduced in \cite{Zhou2022CompositionawareGL}. 
Readability and visual balance mean that when designing posters, designers tend to place text without underlays in a relatively flat area. 
$R_{sub}$ and $R_{shm}$ can reflect the degree of occlusion of key subjects, the lower the better. $R_{occ}$ means the ratio of non-empty layouts predicted by models.

Graphic measures use the same indicators as in \cite{Zhou2022CompositionawareGL}, such as alignment $R_{ali}$, overlap $R_{ove}$ and $R_{und}$. $R_{ove}$ excludes underlays and embellishments, because these two elements are generally attached to other types of elements. At the same time, redefine $R_{und}$ to evaluate the influence of substrate elements on the layout quality. $R_{und}$ and layout quality show a positive correlation.

\subsection{Comparison with Content-Aware Methods}
As mentioned in the previous chapters, ContentGAN and CGL-GAN are two generators considering the influence of image content on layout, so here is our main comparison model. We re-implement ContentGAN based on the released codes\footnote{\href{https://xtqiao.com/projects/content aware layout}{https://xtqiao.com/projects/content aware layout}}, and specifically add content feature extraction and text feature extraction modules consistent with our method. Meanwhile, we tried our best to re-implement the CGL-GAN method based on the details in the paper. The quantitative comparison results of the three methods are shown in Tab. ~\ref{table1}. No matter whether in user study or composition-relevant metric, our method is obviously winning, which shows that the proposed method has a better ability to represent the relationship between image content and layout.

The qualitative evaluation results of different models are shown in Fig.~\ref{fig:main_result}. The three columns on the left show that our model has a stronger subject representation ability, which can effectively highlight the subjects in posters such as commodities and models compared with other methods. From the results in the middle part, due to the introduction of the Visual-Textual Relation-Aware Module (VTRAM), the model can learn where the text should be placed to ensure the text readability and visual balance of the poster layout. The right part shows that our model can also strongly express the relationship between graph elements under the premise of ensuring that the products are not occluded.

\subsection{Comparison with Content-Agnostic Methods}
Similarly, we also compare our model performance with recent content-agnostic SOTA methods \cite{inoue2023layout, BLT2022ECCV}. Based on the released code\footnote{\href{https://github.com/CyberAgentAILab/layout-dm}{https://github.com/CyberAgentAILab/layout-dm}} \footnote{\href{https://shawnkx.github.io/blt}{https://shawnkx.github.io/blt}}, we re-implement the above methods. As shown in Tab. ~\ref{table2}, our model has great advantages in user study and composition-relevant because of the modeling relationship between image content and layout. But is less effective on graphic metrics. We attribute this to the fact that our model needs to consider image content information when generating layouts, such as considering visual balance factors or avoiding the main product area, etc. For the $R_{und}$, although our model does not exceed BLT, it is better than LayoutDM. Because of the introduction of the GRAM, the model learns the relationship between Underlay and other types of layout elements. As shown in the right part in Fig. ~\ref{fig:main_result}, our model is more harmonious in the collocation of text and substrate.
\input{Table/table2}

\subsection{Controllable Layout Generation}
Our model can achieve controllable layout generation, which is also a highlight of our method. We show the layout results of the model under different constraints, which are (1) Text number and content; (2) Given partial layout. 

\textbf{Text number and content.} 
\input{Figtex/ablation_text_num}
\input{Figtex/ablation_text_content}
As shown in Fig.\ref{fig:text_num}, the last three columns represent the layout results of the same background image under different text number constraints. Interestingly, we find that the number of text elements in the layout result is consistent with the number of input text, which proves that our model has learned the relationship between the number of texts and layout elements. 
As shown in Fig. ~\ref{fig:text_content}, the left column indicates that given different text lengths, our method can generate boxes in the appropriate proportion, the right column represents the position of the element affected by the text content. It proves that the proposed model has a sufficient expression between literal semantic information and layout output.
\input{Figtex/user_con}

\textbf{Given partial layout.}
In order to verify whether the output results of the model are acceptable given the part layout, we conduct different experiments and the results are shown in Fig. ~\ref{fig:user_con}. Our model can give qualified results, especially in the results of the third column, our model will not generate additional layouts without enough layout space, which shows that the model has strong constraints and generalization ability.

\subsection{Ablation Studies}
We conduct comparative experiments in the visual-textual relation-aware module, geometry relation-aware module, as well as the layout diversity and rationality.
\input{Table/table3}

\textbf{Visual-Textual Relation-Aware Module.} In order to verify the influence of visual and text attention features on the layout effect, we conduct ablation experiments. Specifically, we train two versions of the model on the same training data: (a) the model contains all modules; (b) the model removes VTRAM. The results can be seen in Tab. ~\ref{table3}. Due to the introduction of the text and image attention mechanism, the model has learned content information related to the composition of the image, which greatly improves the composition-relevant metrics without sacrificing the effectiveness of graph metrics to a certain extent. We believe that multi-modal deep semantic features have a more accurate expression for layout elements.
\input{Table/table4}

\textbf{Geometry Relation-Aware Module.} Geometry Relation-Aware Module (GRAM) is to obtain more robust and accurate box coordinates and sizes after the diffusion process. We remove the GRAM from the proposed model as a ablation comparison model. As shown in Tab. ~\ref{table4}, the model with GRAM has a 0.4$\%$ reduction on $R_{ali}$, a 0.07$\%$ improvement on $R_{und}$ and a 3.7$\%$ reduction on $R_{ove}$, which is attributed to the more accurate description of the boxes in the process of generating the layout. In particular, the performance of composition-relevant metrics has also been improved, because the influence of image information on the position of elements is also considered in the introduction of GRAM. In general, GRAM can achieve a balance in the improvement of composition-relevant metrics and graphic metrics.

\textbf{Layout diversity and rationality.} 
\input{Figtex/ablation_diversity}
Because our method will give some random layout boxes at the beginning of the inference stage, in order to evaluate the layout diversity and rationality of the model, we give qualitative experimental results. From left to right, Fig. \ref{fig:diversity} shows the layout results corresponding to five different layouts by random seeds at the beginning of inference. From top to bottom, Fig. \ref{fig:diversity} also shows the layout results of different images under the same random seed. Although the resulting layout results are different, they are all reasonable, indicating the diversity and rationality of the layout model.

\section{Conclusion}
In this paper, we propose a relation-aware diffusion model to generate poster layouts, in which the relationship between visual and textual contents and the relationship between elements are considered to help get pleasant layouts. To better integrate visual and textual features, we design a Visual-Textual Relation-Aware Module (VTRAM) to learn the relationship between visual and textual contents. As the coordination of element positions is important for layout, a Geometry Relation-Aware Module (GRAM) is employed to enhance features based on the relative position between elements. In addition, we build a large poster layout dataset, named CGL-Dataset V2. We conduct extensive experiments to prove that the proposed method significantly outperforms the existing methods and can achieve controllable generation. Ablation studies also demonstrate the effectiveness of VTRAM and GRAM.
\vfill \eject
\bibliographystyle{ACM-Reference-Format}
\balance
\bibliography{sample-base}

\end{document}

%% file: Figtex/compare1.tex
\begin{figure}[t] 
\centering 
\includegraphics[width=0.9\linewidth]{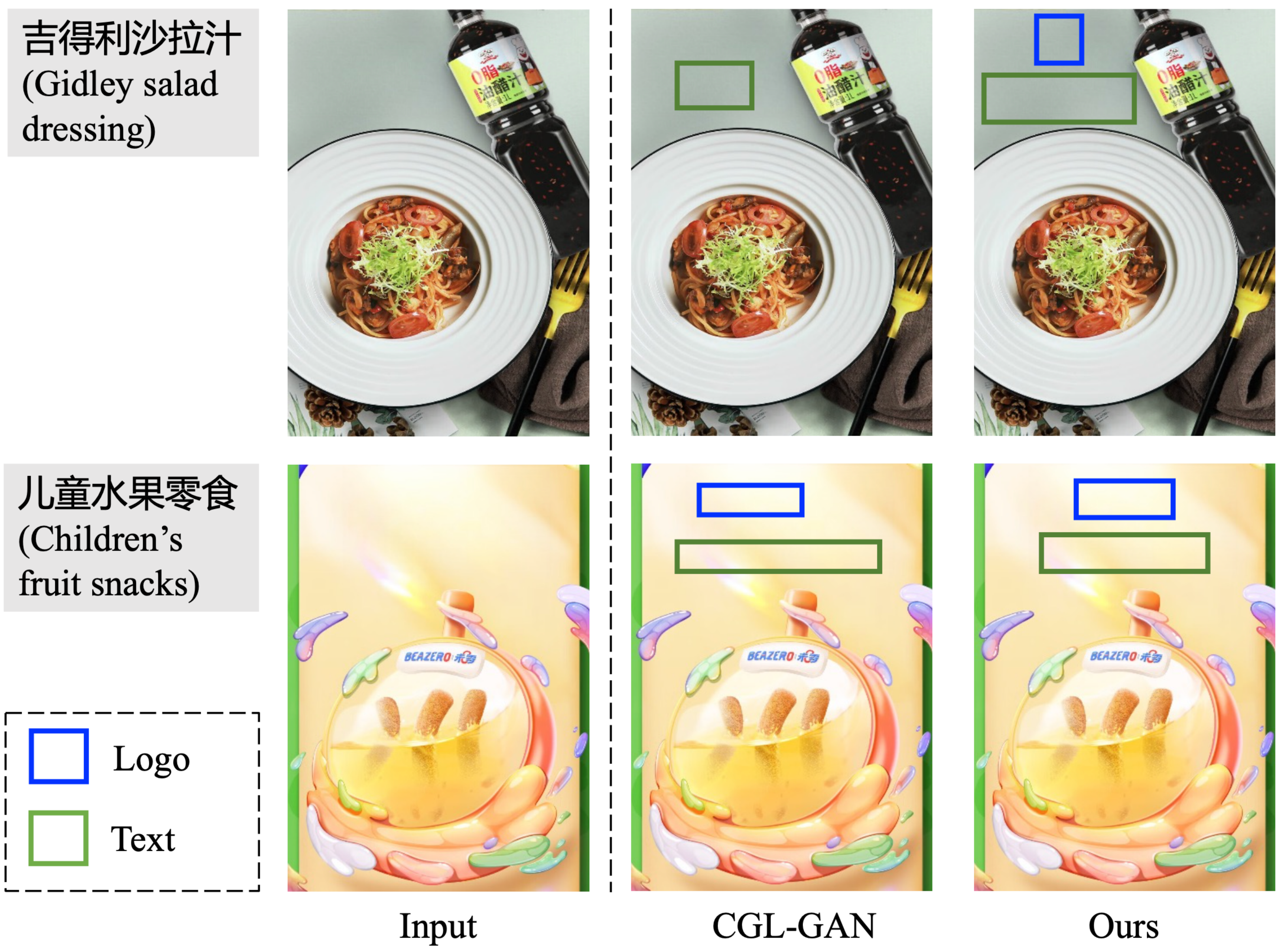} 
\vspace{-1em}
\caption{The visual examples of poster layout produced by CGL-GAN\cite{Zhou2022CompositionawareGL} and ours.}
\label{fig:compare1}
\end{figure}

%% file: Figtex/dataset.tex
\begin{figure*}[ht] 
\centering 
\includegraphics[width=0.9\linewidth]{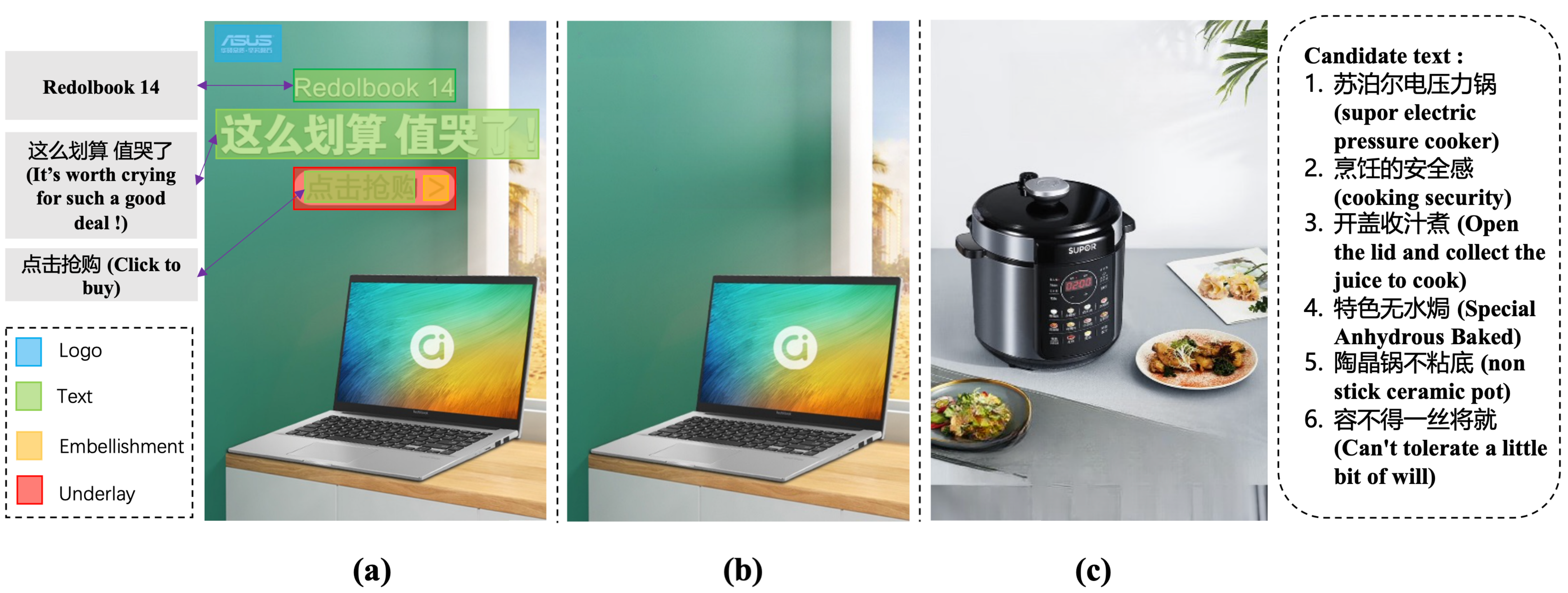} 
\vspace{-1em}
\caption{(a) Poster layout annotation. Different colors represent different element types, the text annotation results are in the gray box, and the English translation is in brackets; (b) Clean image; (c) Input for inference stage.}
\label{fig:dataset}
\end{figure*}

%% file: Figtex/overview.tex
\begin{figure*}[ht] 
\centering 
\includegraphics[width=0.8\linewidth]{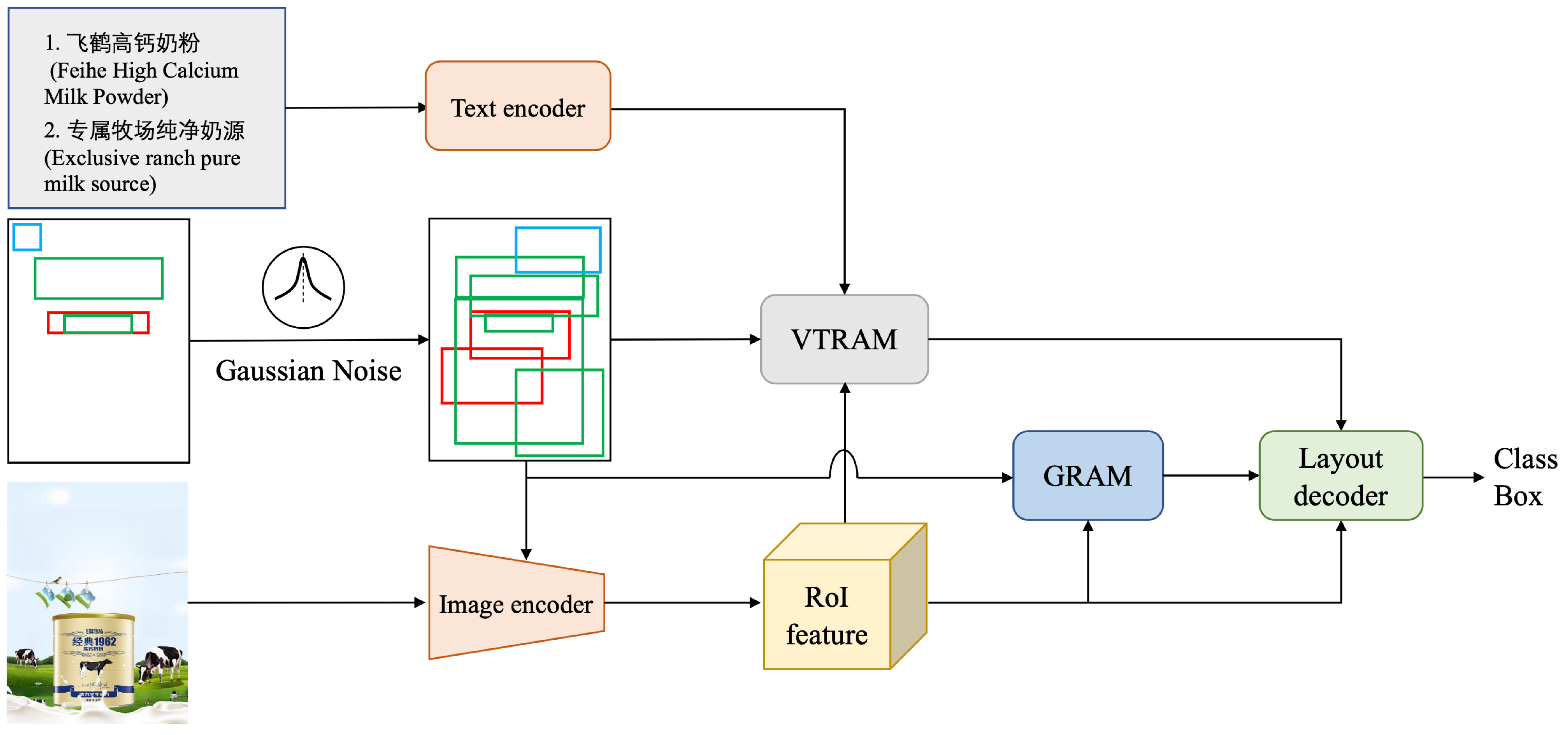} 
\vspace{-1em}
\caption{The overview of our method, which contains four parts: feature extractor, VTRAM, GRAM and layout decoder.}
\label{fig:overview}
\end{figure*}

%% file: Figtex/diffusion.tex
\begin{figure}[ht] 
\centering 
\includegraphics[width=1\linewidth]{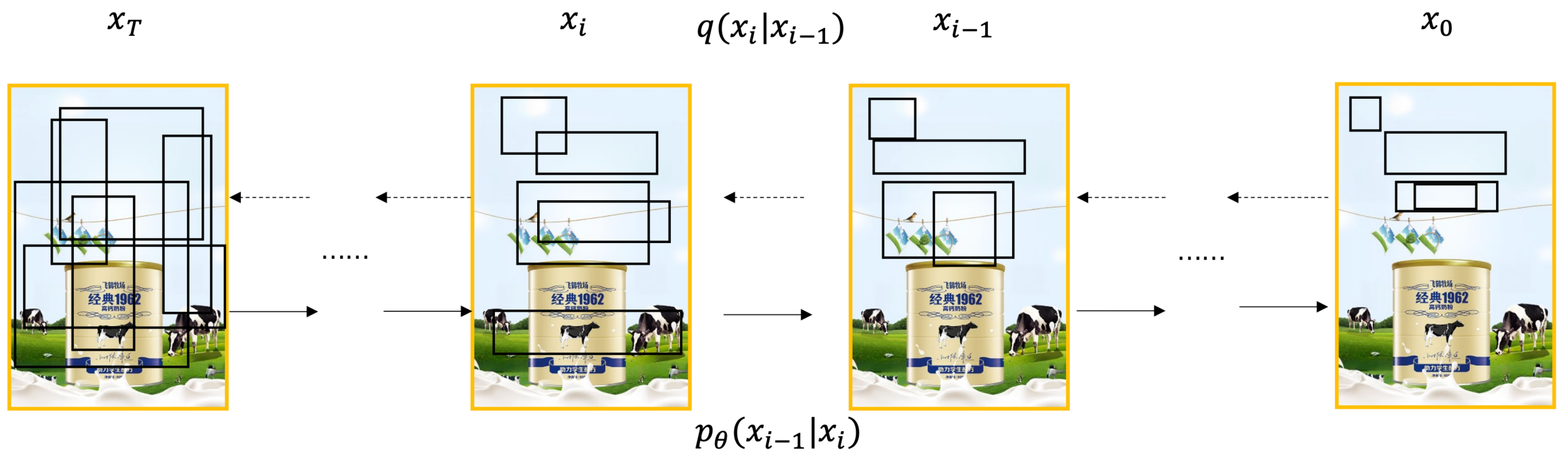} 
\vspace{-1em}
\caption{Inspired by diffusion denoising process, from left to right, we formulate the poster layout generation as a process to gradually refine the position and size of boxes from step $T$ to step $i$. }
\label{fig:diffusion}
\end{figure}

%% file: Figtex/VTRAM.tex
\begin{figure}[ht] 
\centering 
\includegraphics[width=0.9\linewidth]{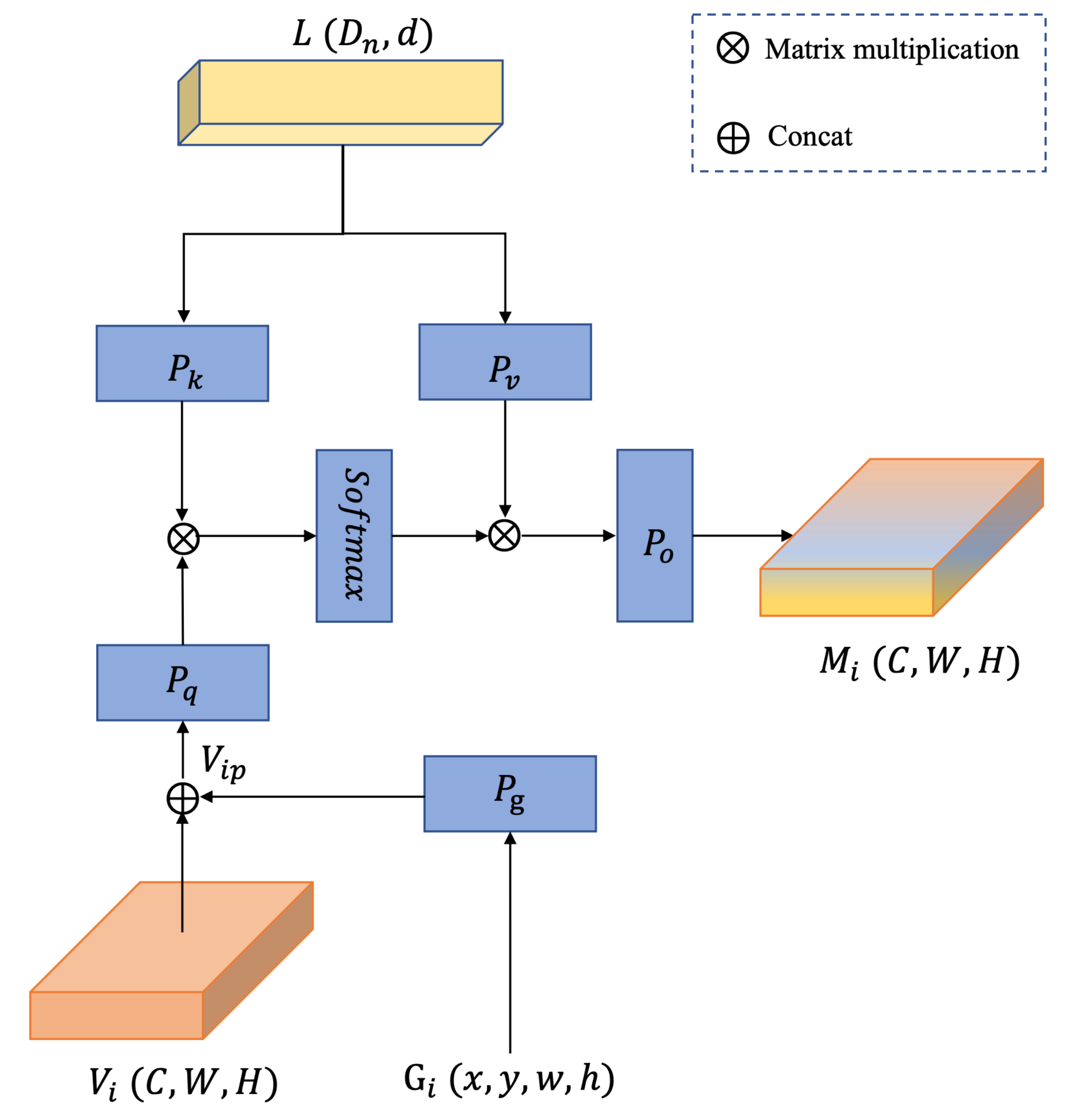} 
\caption{The overview of the VTRAM. 
As illustrated in the figure, it takes as input text features, RoI features and corresponding coordinates.
The coordinate information is first embedded into RoI features to get $V_{ip}$.
Next, the scaled dot-product attention\cite{Vaswani2017AttentionIA} is calculated using the visual position feature $V_{ip}$ as the query, and text features $L$ as the key and value.}
\label{fig:VTRAM}
\end{figure}

%% file: Figtex/GRAM.tex
\begin{figure}[ht] 
\centering 
\includegraphics[width=0.9\linewidth]{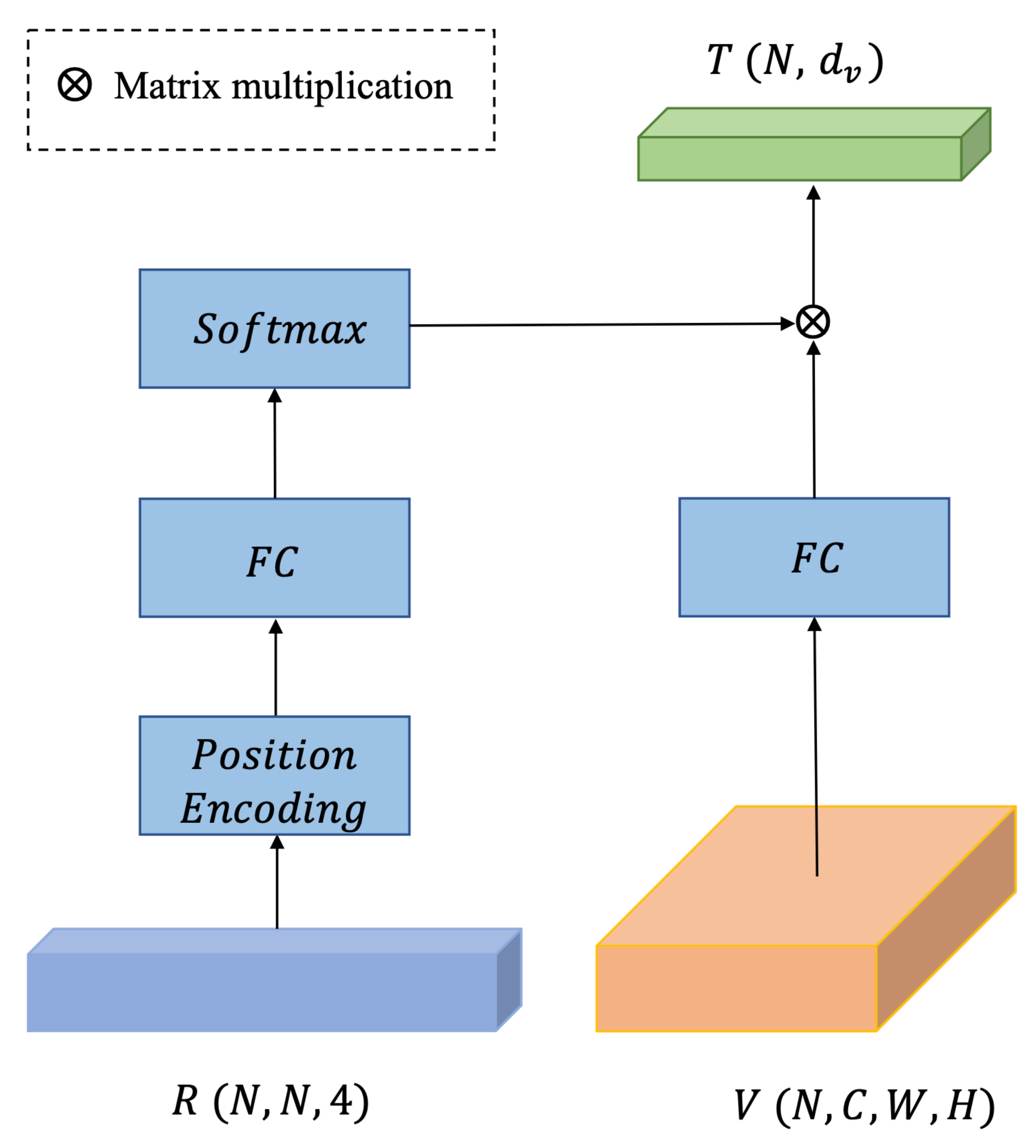} 
\caption{The overview of GRAM. 
It exploits the relative positional relationships between elements. The input consists of two parts: relative position features $R$ and RoI features $V$.}
\label{fig:GRAM}
\end{figure}

%% file: Figtex/main_result.tex
\begin{figure*}[ht] 
\centering 
\includegraphics[width=0.85\linewidth]{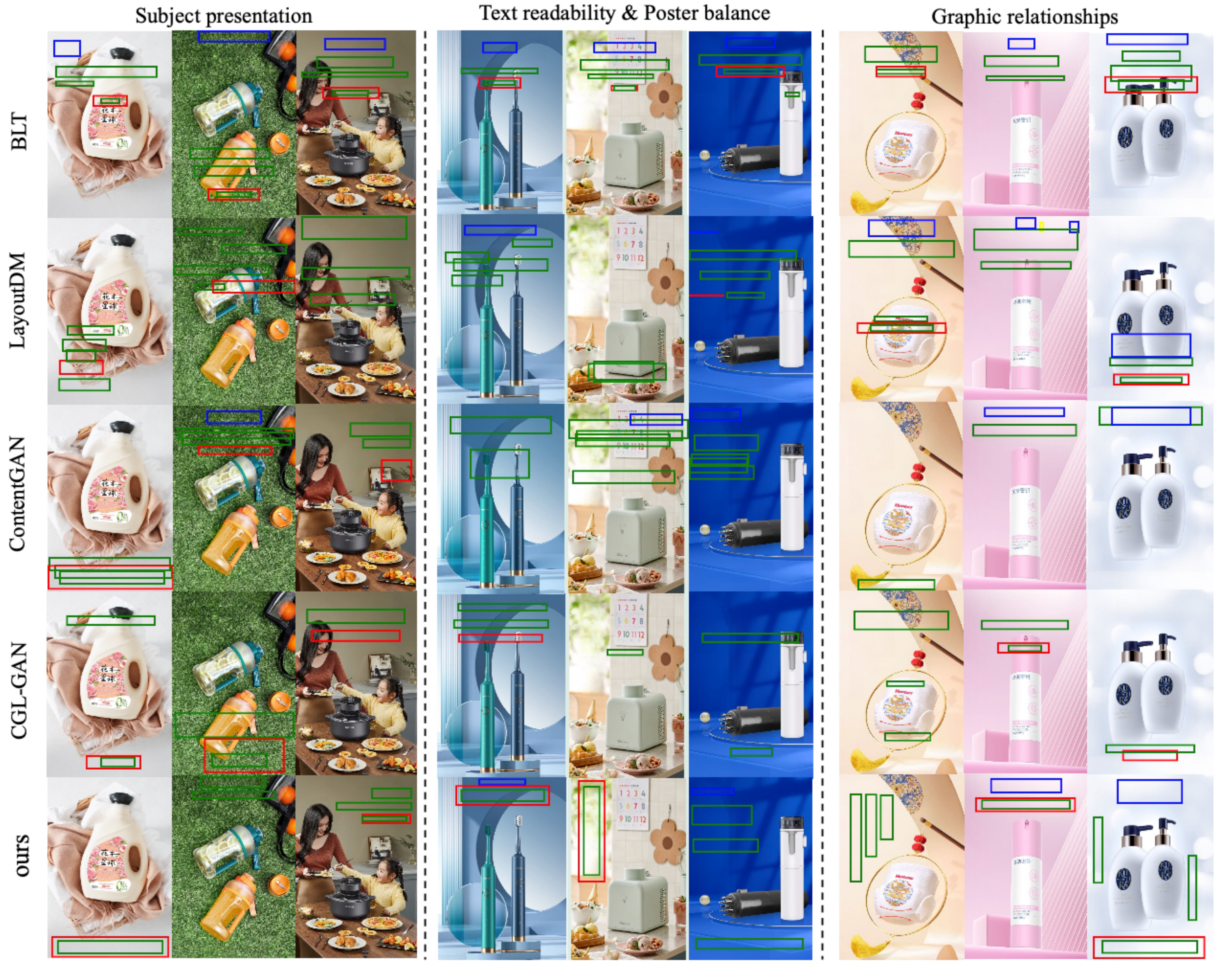} 
\vspace{-1em}
\caption{Qualitative comparison results with SOTA methods. Each column layout represents the results obtained by different methods for the same image, and each row represents the layout results of the same method for different images.}
\label{fig:main_result}
\end{figure*}

%% file: Table/table1.tex
\begin{table*}[t]
\caption{Comparison with content-aware methods.}
\resizebox{\linewidth}{!}{
\begin{tabular}{c|cccc|cccc|ccc}
\hline
\multirow{2}{*}{Model} & \multicolumn{4}{c|}{User study} & \multicolumn{4}{c|}{Composition-relevant measures} & \multicolumn{3}{c}{Graphic measures} \\ \cline{2-12} 
           & $P^{*}_{qs} \uparrow $ & $P^{*}_{best} \uparrow$  & $P_{qs}\uparrow$  & $P_{best}\uparrow$  & $P_{shm}\downarrow$  & $P_{com}\downarrow$  & $P_{sub}\downarrow$  & $P_{occ}\uparrow$  & $P_{ali}\downarrow$  & $P_{ove}\downarrow $ & $P_{und}\uparrow $ \\ \hline
ContentGAN & 26.1\% & 12.8\% & 30.6\% & 7.2\% & 23.610 & 31.930 & 0.767 & \textbf{1.000} & 0.009 & 0.065 & 0.840 \\
CGL-GAN    & 28.3\% & 16.1\% & 44.4\% & 8.9\% & 21.670 & 16.040  & 0.772 & 0.875 & \textbf{0.007} & 0.081 & 0.732 \\
Ours       & \textbf{75.6\%} & \textbf{66.7\%} & \textbf{86.7\%} & \textbf{78.9\%} & \textbf{15.970} & \textbf{10.260}  & \textbf{0.742} & 0.997 & 0.008 & \textbf{0.046} & \textbf{0.983} \\
\hline
\end{tabular}}
\label{table1}
\end{table*}

%% file: Table/table2.tex

\begin{table*}[t]
\caption{Comparison with content-agnostic methods.}
\resizebox{\linewidth}{!}{
\begin{tabular}{c|cccc|cccc|ccc}
\hline
\multirow{2}{*}{Model} & \multicolumn{4}{c|}{User study} & \multicolumn{4}{c|}{Composition-relevant measures} & \multicolumn{3}{c}{Graphic measures} \\ \cline{2-12} 
           & $P^{*}_{qs} \uparrow $ & $P^{*}_{best} \uparrow$  & $P_{qs}\uparrow$  & $P_{best}\uparrow$  & $P_{shm}\downarrow$  & $P_{com}\downarrow$  & $P_{sub}\downarrow$  & $P_{occ}\uparrow$  & $P_{ali}\downarrow$  & $P_{ove}\downarrow $ & $P_{und}\uparrow $ \\ \hline
BLT & 57.2\% & 21.6\% &57.8\% & 26.1\% & 22.450 & 28.540 & 0.765 & 1.000 & \textbf{0.004} & \textbf{0.002} & \textbf{0.993} \\
LayoutDM & 32.8\% & 13.8\% &37.2\% & 22.8\% & 21.300 & 34.310 & 0.763 & \textbf{1.000} & 0.006 & 0.039 & 0.896 \\
Ours & \textbf{75.6\%} & \textbf{58.9\%} &\textbf{82.2\%} & \textbf{46.7\%} & \textbf{15.970} & \textbf{10.260} & \textbf{0.742} & 0.997 & 0.008 & 0.046 & 0.983 \\
\hline 
\end{tabular}
}
\label{table2}
\end{table*}

%% file: Figtex/ablation_text_num.tex
\begin{figure}[t] 
\centering 
\includegraphics[width=0.9\linewidth]{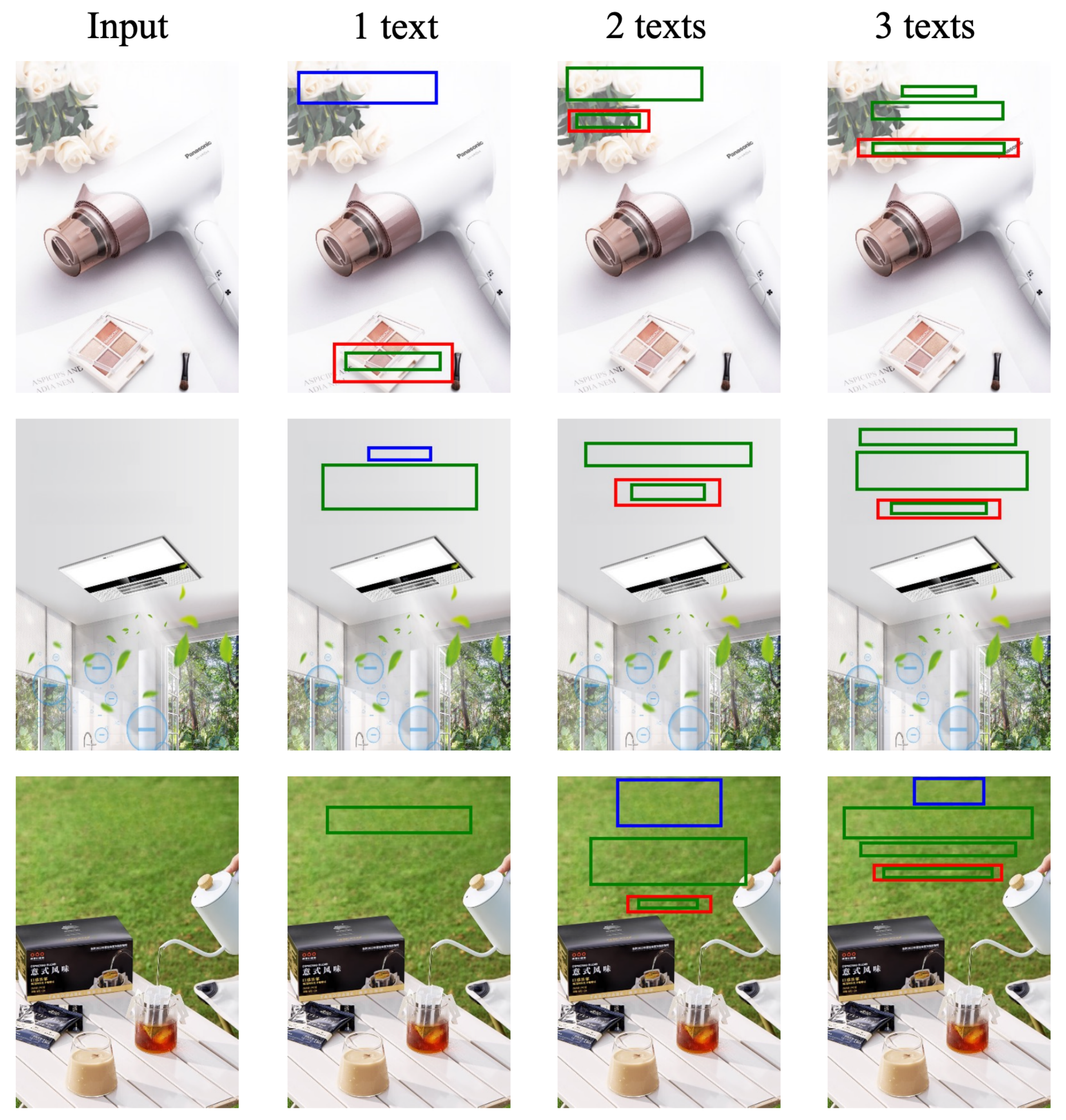} 
\vspace{-1em}
\caption{Layout results with different amounts of text. The second to fourth columns represent a range of 1 to 3 input texts, respectively.}
\label{fig:text_num}
\end{figure}

%% file: Figtex/ablation_text_content.tex
\begin{figure}[t] 
\centering 
\includegraphics[width=1\linewidth]{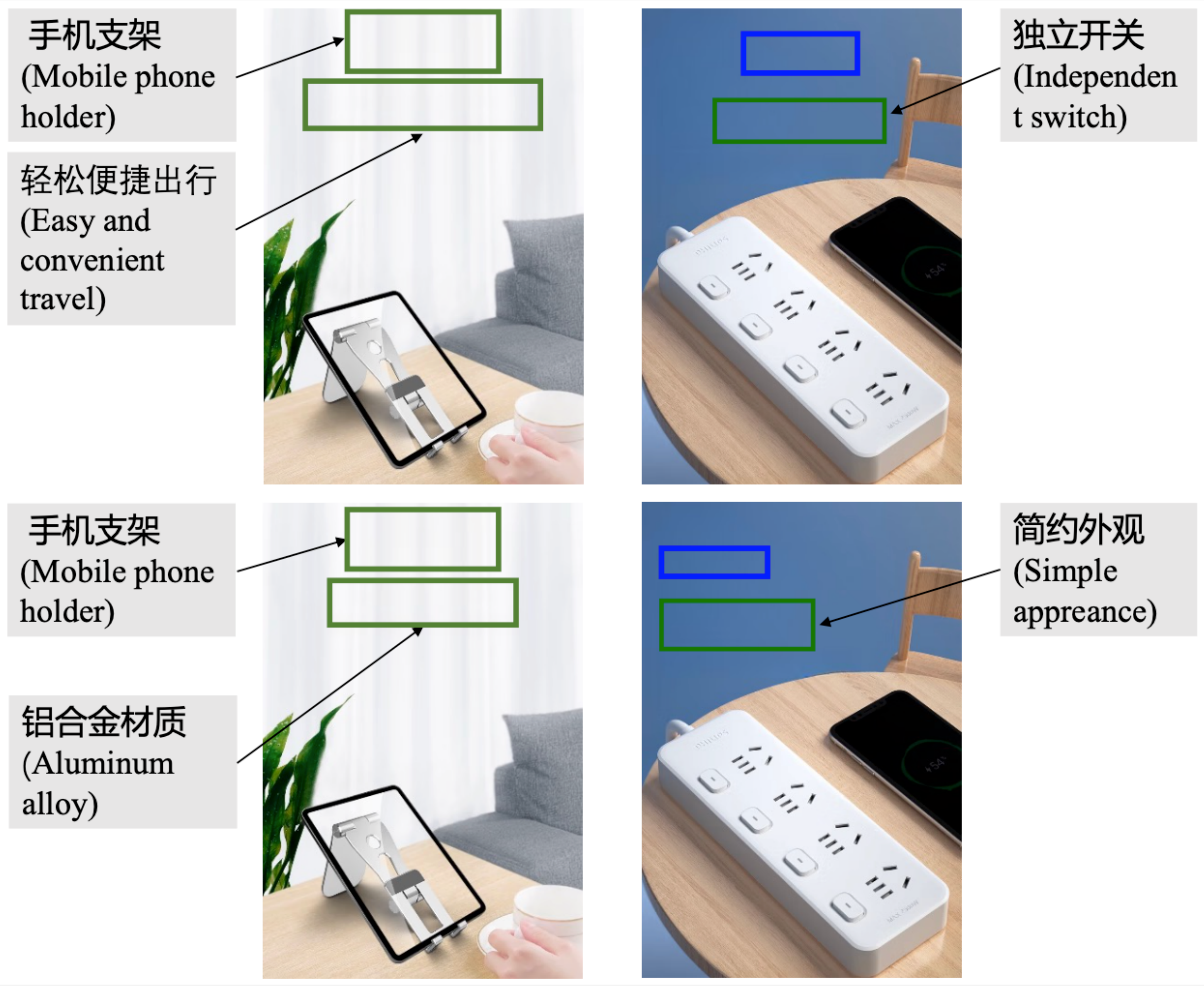} 
\vspace{-1em}
\caption{Layout results with different text lengths (left column) and contents (right column).}
\label{fig:text_content}
\end{figure}

%% file: Figtex/user_con.tex
\begin{figure}[t] 
\centering 
\includegraphics[width=1\linewidth]{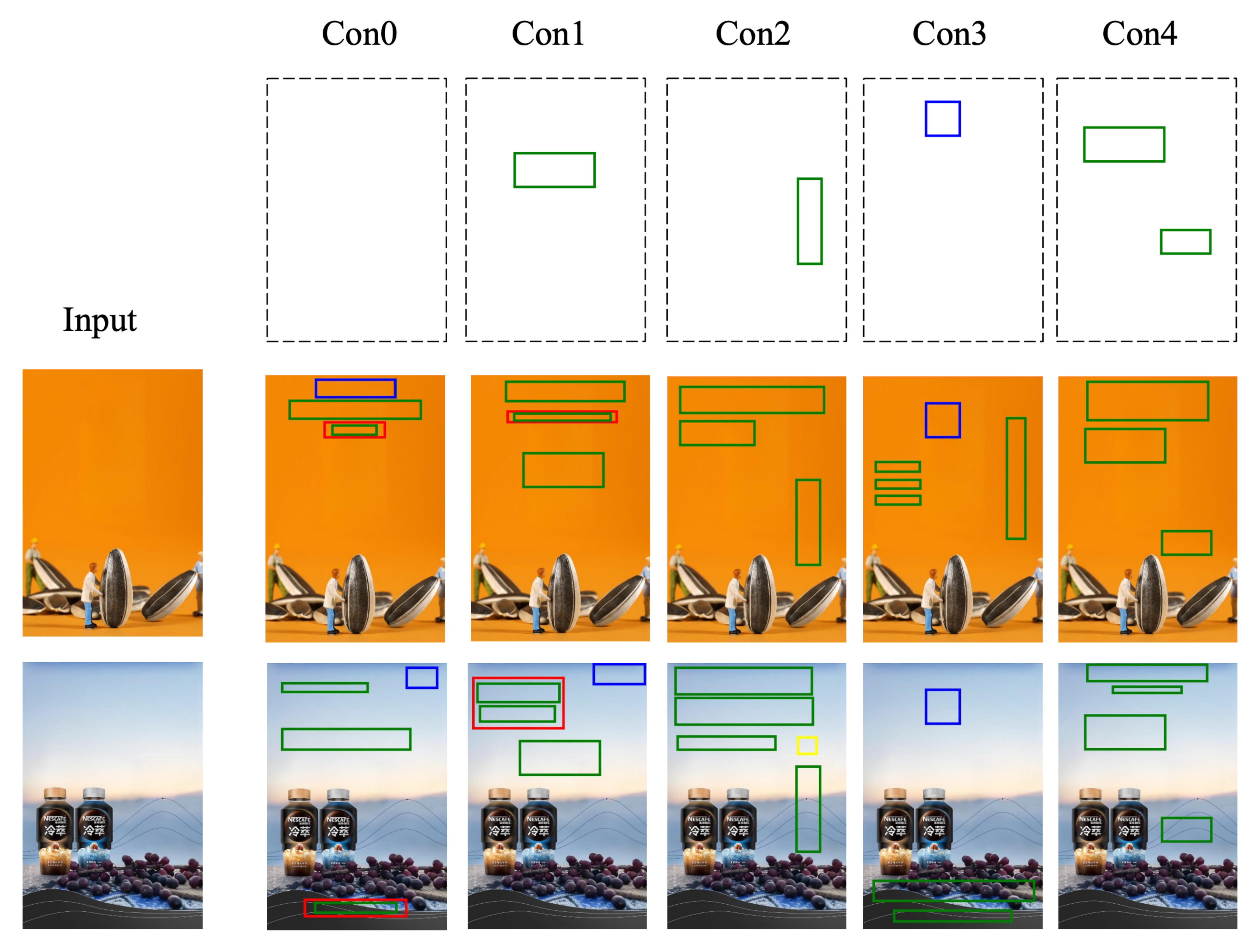} 
\vspace{-1em}
\caption{Layout results under different user constraints.}
\label{fig:user_con}
\end{figure}

%% file: Table/table3.tex
\begin{table}[t]
\caption{Ablation studies of VTRAM. Ours$^{*}$ means our model without VTRAM.}
\resizebox{\linewidth}{!}{
\begin{tabular}{c|cccc|ccc}
\hline
Model   & $P_{shm}\downarrow$  & $P_{com}\downarrow$  & $P_{sub}\downarrow$  & $P_{occ}\uparrow$  & $P_{ali}\downarrow$  & $P_{ove}\downarrow $ & $P_{und}\uparrow $ \\ \hline
Ours$^{*}$ & 17.450 & 12.720 & 0.764 & 0.989 & 0.010 & 0.053 & \textbf{0.987} \\
Ours    &\textbf{15.970} & \textbf{10.260}  & \textbf{0.742} & \textbf{0.997} & \textbf {0.008} & \textbf{0.046} & 0.983 \\ \hline
\end{tabular}}
\label{table3}
\end{table}

%% file: Table/table4.tex

\begin{table}[t]
\caption{Ablation studies of GRAM. Ours$^{*}$ means our model without GRAM.}
\resizebox{\linewidth}{!}{
\begin{tabular}{c|cccc|ccc}
\hline
Model   & $P_{shm}\downarrow$  & $P_{com}\downarrow$  & $P_{sub}\downarrow$  & $P_{occ}\uparrow$  & $P_{ali}\downarrow$  & $P_{ove}\downarrow $ & $P_{und}\uparrow $ \\ \hline
Ours$^{*}$ & 17.190 & \textbf{10.120} & 0.753 & 0.922 & 0.012 & 0.083 & 0.976 \\
Ours    & \textbf{15.970} & 10.260  & \textbf{0.742} & \textbf{0.997} & \textbf{0.008} & \textbf{0.046} & {\textbf{0.983}} \\ \hline
\end{tabular}}
\label{table4}
\end{table}

%% file: Figtex/ablation_diversity.tex
\begin{figure}[t] 
\centering 
\includegraphics[width=1\linewidth]{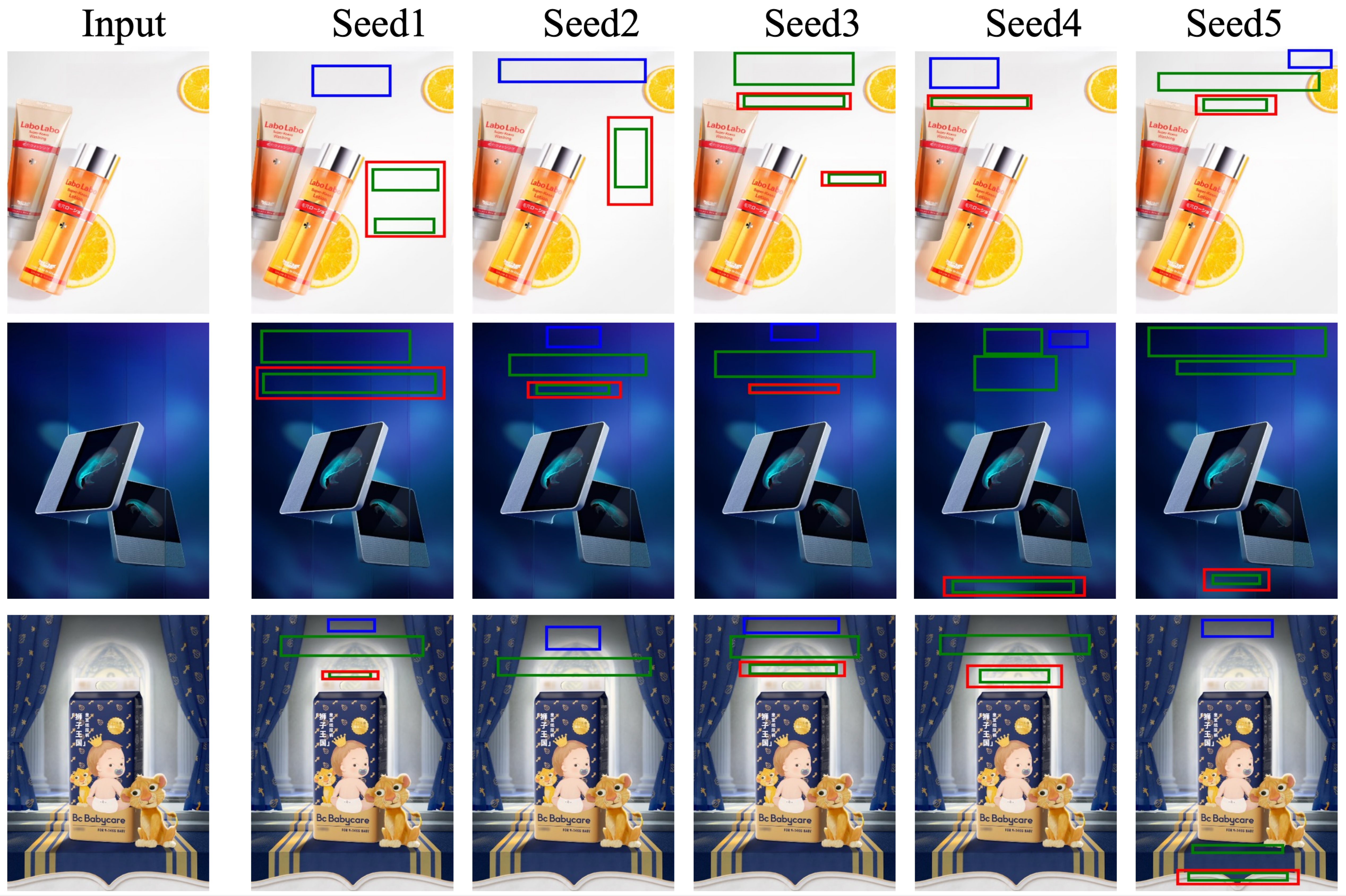} 
\vspace{-1em}
\caption{Generated layouts under different random seeds. Each row is the result of the same input image under different random seeds, and each column the different images under the same random seed.}
\label{fig:diversity}
\end{figure}